# Qualitative MDPs and POMDPs: An Order-of-Magnitude Approximation

Blai Bonet and Judea Pearl
Cognitive Systems Laboratory
Department of Computer Science
University of California, Los Angeles
Los Angeles, CA 90024
{bonet,judea}@cs.ucla.edu

## Abstract

We develop a qualitative theory of Markov Decision Processes (MDPs) and Partially Observable MDPs that can be used to model sequential decision making tasks when only qualitative information is available. Our approach is based upon an order-of-magnitude approximation of both probabilities and utilities, similar to $\epsilon$-semantics. The result is a qualitative theory that has close ties with the standard maximum-expected-utility theory and is amenable to general planning techniques.

## 1 Introduction

The general task of sequential decision making under uncertainty and partial information is of central importance in AI since it embraces a broad range of common problems found in planning, robotics, game solving, etc. Currently, the most general and clear formulation of the task is achieved through the theory of Markov Decision Processes (MDPs) and Partially Observable MDPs (POMDPs) [8, 16, 4]. These models do not only provide a sound, concise and general framework for modeling complex problems but also algorithms for solving them, the most important being the Value Iteration and Policy Iteration algorithms.

The standard formulation for an MDP (or POMDP) consists of two types of ingredients:

(a) qualitative information that define the *structure* of the problem, e.g. the set of world configurations (state space), the set of available decisions (also known as controls or actions), the feedback that can be received by the agent, etc. and

(b) quantitative information (also known as parametrization of the structure) that, together with the qualitative information, defines the model. Examples of the quantitative information are the transition probabilities of going from one state to another after the application of a control, the costs incurred in such application, etc.

In general, the (optimal) solution to an MDP (POMDP) depends in both types of information, so the standard algorithms need such information. Quite often, however, we have precise knowledge of the qualitative information but only "rough" estimates of the quantitative parameters. In such cases, the standard algorithms cannot be applied unless the missing information is "completed" – a process that is often arbitrary and unnecessary for obtaining reasonable solutions. Ideally, we would like to have a well-founded framework in which partially specified problems can be solved.[1]

In this paper, we develop a *qualitative theory* for MDPs and POMDPs in which such underspecified problems can be modeled and solved. As it will be seen, the resulting theory can be thought as a generalization of the standard MDP theory in the sense that as the quantitative information becomes more precise, the qualitative processes become "closer" to the standard processes.

More precisely, we will show how a qualitative description of the sequential decision task using kappa rankings can be translated into a "consistent" description using polynomials in a dummy variable $\varepsilon$. Then, standard algorithms like Value Iteration can be applied for obtaining optimal solutions. The reason for doing this translation and not performing Value Iteration over the input description is that order-of-magnitude quantities do not increase through summations, so the output from Value Iteration cannot be used to discriminate among policies. As it will be seen, the contributions of this paper are three: (*i*) it gives a formal foundation for a sequential decision theory based on poly-nomials in $\varepsilon$, $(ii)$ shows how to consistently translate descriptions of tasks using kappa rankings into models using a more expressive language, and $(iii)$ shows how standard planning algorithms can be applied to solve the resulting models.

---
[1] Other approaches for solving unspecified MDPs are those based on Reinforcement Learning, yet they basically estimate the quantitative information from experimentation; see [26].

The paper is organized as follows. In the next Sect. we review the formal definitions and most important results of the standard theory of MDPs and POMDPs. In Sect. 3, we present the formal foundations upon which the qualitative theory of MDPs and POMDPs is built. Sects. 4 and 5 present the qualitative theory of MDPs and POMDPs, while Sect. 6 discuss some computational issues that arise from special subclasses of tasks. We finish the paper with a brief discussion that includes related work and a summary. Due to space limitations, we only provide proofs for the most novel results.

## 2 Standard MDPs and POMDPs

In this section and the rest of the paper we use a notation similar to the one in [3], the reader is referred there for an excellent introduction to MDPs.

The MDP model assumes the existence of a physical system that evolves in discrete time and that is controlled by an agent. The system dynamics is governed by probabilistic transition functions that map states and controls to states. At every time, the agent incurs in a cost that depends in the current state of the system and the applied control. Thus, the task is to find a control strategy (also known as a policy) that minimize the expected total cost over the infinite horizon time setting. Formally, an MDP is characterized by

(M1) A finite state space $S = \{1, \ldots, n\}$,

(M2) a finite set of controls $U(i)$ for each state $i \in S$,

(M3) transition probabilities $p_{i,u}(j)$ for all $u \in U(i)$ that are equal to the probability of the next state being $j$ after applying control $u$ in state $i$, and

(M4) a cost $g(i, u)$ associated to $u \in U(i)$ and $i \in S$.

A strategy or policy $\pi$ is an infinite sequence $(\mu_0, \mu_1, \ldots)$ of functions where $\mu_k$ maps states to controls so that the agent applies the control $\mu_k(i)$ in state $x_k = i$ at time $k$, the only restriction being that $\mu_k(i) \in U(i)$ for all $i \in S$. If $\pi = (\mu, \mu, \ldots)$, the policy is called *stationary* (i.e. the control does not depend in time) and is simply denoted by $\mu$. The cost associated to $\pi$ when the system starts at state $x_0$ is:

$$J_\pi(x_0) \doteq \lim_{N \to \infty} E \left\{ \sum_{k=0}^{N-1} \alpha^k g(x_k, \mu_k(x_k)) \right\} \quad (1)$$

where the expectation is taken with respect to the probability distribution induced by the transition probabilities, and where the number $\alpha \in [0, 1]$, called the *discount factor*, is used to discount future costs at a geometric rate.

The MDP *problem* is to find an *optimal policy* $\pi^*$ satisfying $J^*(i) \doteq J_{\pi^*}(i) \leq J_\pi(i)$ $(i = 1, \ldots, n)$, for every other policy $\pi$. Although there could be none or more than one optimal policy, the optimal cost vector $J^*$ is always unique. The case $\alpha < 1$ is of outmost importance since it guarantees that the optimal policy always exists and, more important, that there exists a stationary policy that is optimal. In such case, $J^*$ is the unique solution to the *Bellman* equation:

$$J^*(i) = \min_{u \in U(i)} \left\{ g(i, u) + \alpha \sum_{j=1}^{n} p_{i,u}(j) J^*(j) \right\}. \quad (2)$$

Also, if $J^*$ is a solution for (2) then the *greedy* stationary policy $\mu^*$ with respect to $J^*$:

$$\mu^*(i) \doteq \operatorname*{argmin}_{u \in U(i)} \left\{ g(i, u) + \alpha \sum_{j=1}^{n} p_{i,u}(j) J^*(j) \right\} \quad (3)$$

is an optimal and stationary policy for the MDP. Therefore, solving the MDP is equivalent to solving (2). Such equation can be solved using the DP *operator*:

$$(TJ)(i) \doteq \min_{u \in U(i)} \left\{ g(i, u) + \alpha \sum_{j=1}^{n} p_{i,u}(j) J(j) \right\} \quad (4)$$

that maps $\mathbb{R}^n$ into $\mathbb{R}^n$. When $\alpha < 1$ the DP operator is a contraction mapping with fixed point $J^*$ that satisfy:[2]

$$J^* = TJ^* = \lim_{k \to \infty} T^k J_0 \quad (5)$$

where $J_0$ is any $n$-dimensional vector. The Value Iteration algorithm computes $J^*$ iteratively by using (2) as an update rule. Starting from any vector $J$, Value Iteration computes a succession of vectors $\langle J_k \rangle_{k \geq 0}$ defined by $J_0 \doteq J$ and $J_{k+1} \doteq TJ_k$. The algorithm stops when $J_{k+1} = J_k$, or when the residual $\max_{i \in S} |J_{k+1}(i) - J_k(i)|$ is sufficiently small. In the latter case, the suboptimality of the resulting policy is bounded by a constant multiplied by the residual.

### Partially Observable MDPs

A Partially Observable Markov Decision Process (POMDP) is an MDP in which the agent does not know the state of the system. This is an important departure from the MDP model since even if the agent knows the optimal strategy for the underlying MDP, it cannot apply it. Thus, the agent needs to estimate the

---

[2] A contraction mapping $T : S \to S$ over a Banach space $S$ with norm $\|\cdot\|$ is a bounded operator such that $\|TJ - TJ'\| \leq \gamma \|J - J'\|$ for some $\gamma < 1$. In this case, it is known that $T$ is continuous, that has a unique fixed point $J^*$, and that $T^n J \to J^*$ as $n \to \infty$ for any $J \in S$. See [18].

state of the system and then act accordingly. Such estimates are known as the *belief states* of the agent and are updated continuously as the system evolves. The POMDP framework also extends the MDP framework by allowing controls to return information about the system. For example, a `blood-test` might return blood type and `reading-radar` might return the distance to objects. See [6] for definitions from the AI perspective. Formally, a POMDP is characterized by:

(P1) A finite state space $S = \{1, \ldots, n\}$,

(P2) a finite set of controls $U(i)$ for each state $i \in S$,

(P3) transition probabilities $p_{i,u}(j)$ for all $u \in U(i)$ equal to the probability of the next state being $j$ after applying $u$ in $i$,

(P4) a finite set of observations $O(i, u) \subseteq O$ that may result after applying $u \in U(i)$ in $i \in S$,

(P5) observation probabilities $p_{i,u}(o)$ for all $u \in U(i)$ and $o \in O(i, u)$ equal to the probability of receiving $o$ in $i$ after applying $u$, and

(P6) a cost $g(i, u)$ associated to $u \in U(i)$ and $i \in S$.

It had been shown that finding an optimal strategy to this problem is equivalent to solving an *infinite-state* MDP problem in *belief space*, the so-called belief-MDP, whose elements are:

(B1) A belief space $B$ of prob. distributions over $S$,

(B2) a set of controls $U(x) = \{u : \forall i[x(i) > 0 \Rightarrow u \in U(i)]\}$ for each belief state $x \in B$, and

(B3) a cost $g(x, u) = \sum_{i=1}^{n} g(i, u) x(i)$ for each $u \in U(x)$ and $x \in B$.

The transition probabilities of the belief-MDP are determined by the abilities of the agent. It is known that a full capable and rational agent should perform *Bayesian* updating in order to behave optimally. In that case, the transition probabilities are

$$x \rightsquigarrow x_u^o \quad \text{with probability} \quad p_{x,u}(o)$$

where $u \in U(x)$, $o \in O(x, u)$ the set of possible observations after applying control $u$ in belief state $x$, $p_{x,u}(o)$ is the probability of receiving observation $o$ after applying $u$ in $x$, and $x_u^o$ is the Bayesian update of $x$ after $u$ and $o$; i.e.

$$x_u^o(i) \doteq \frac{x_u(i) \, p_{i,u}(o)}{p_{x,u}(o)}, \tag{6}$$

$$x_u(i) \doteq \sum_{j=1}^{n} x(j) \, p_{j,u}(i), \tag{7}$$

$$p_{x,u}(o) \doteq \sum_{i=1}^{n} x_u(i) \, p_{i,u}(o), \tag{8}$$

$$O(x, u) \doteq \{o : p_{x,u}(o) > 0\}. \tag{9}$$

The corresponding DP operator is:

$$(TJ)(x) \doteq \min_{u \in U(x)} \left\{ g(x, u) + \alpha \sum_{o \in O(x,u)} p_{x,u}(o) J(x_u^o) \right\}.$$

As in the MDP case, when $\alpha < 1$, the DP operator is a contraction mapping so it has a unique fixed point that is the solution to the Bellman equations. This fact guarantees the existence of an stationary strategy that is optimal. Unfortunately, the Value and Policy Iteration algorithms are no longer feasible since the state space is infinite, but see [15] for a survey of POMDPs algorithms.

## 3 Foundations

Our approach to a qualitative theory for MDPs and POMDPs is based on the qualitative decision theory proposed by Wilson few years ago [27]. Wilson's theory, built upon ideas of Pearl and Goldszmidt [22, 21, 14], defines a set of abstract quantities called *extended reals*, denoted by $\mathcal{Q}$, that are used to represent qualitative probabilities and utilities. Each extended real is a *rational function* $p/q$ where $p$ and $q$ are polynomials in $\varepsilon$ with rational coefficients. Plainly, $\varepsilon$ is thought as a very small but unknown quantity so that the extended reals are to be interpreted as "information up to $\varepsilon$ precision." For example, quantities like $1 - \varepsilon$ and $\varepsilon$ might be used for qualitative probabilities like "likely" and "unlikely", and $\varepsilon^{-1}$ for a high utility. These quantities are then combined using standard arithmetic operations between polynomials to compute expected qualitative utilities. The resulting utilities rank the different possible scenarios using a linear order $\succeq$ that is defined on $\mathcal{Q}$ [27].[3]

In order to define a qualitative version of MDPs using Wilson's extended reals we need to be sure that equations like (1) and (5) are well-defined. That is, we need to define a notion of convergence and give conditions that guarantee the existence of such limits. We will get these notions by considering not only polynomials but infinite series in $\varepsilon$.

### 3.1 A Complete Extension of $\mathcal{Q}$

Let $\mathcal{S}_\rho$ be the set of *two-sided infinite formal series* in $\varepsilon$ with real coefficients $s = \sum_k a_k \varepsilon^k$ such that

$$\|s\|_\rho \doteq \sum_k |a_k| \rho^{-k} < \infty \tag{10}$$

where $\rho > 0$ the summation is over all integers. We say that a sequence $\langle s_n \rangle_{n \geq 0}$ *converges* to $s \in \mathcal{S}_\rho$ iff $\|s - s_n\|_\rho \to 0$ as $n \to \infty$, and that the sequence is

---

[3]More precisely, Wilson's paper defines $f \succeq g$ iff there exists $\xi > 0$ such that $f(\varepsilon) - g(\varepsilon) \geq 0$ for all $\varepsilon \in (0, \xi)$.

*fundamental* iff $\|s_n - s_m\|_\rho \to 0$ as $n,m \to \infty$. Note that there is no a priori reason for a fundamental sequence to converge to something in $\mathcal{S}_\rho$. Fortunately, this is not the case and such fact is known as that $\langle \mathcal{S}_\rho, \|\cdot\|_\rho \rangle$ is a *complete* normed space (i.e. a Banach space).[4] The following arithmetic operations are defined in the standard way:

$$(s+t)(k) \doteq s(k) + t(k),$$
$$(\alpha s)(k) \doteq \alpha s(k),$$
$$(s \cdot t)(k) \doteq \sum_{i,j} [\![i+j=k]\!]s(i)t(j) = \sum_i s(i)\,t(k-i)$$

where $s(k)$ is the functional notation for the $k$th term of $s$, and $[\![\varphi]\!]$ is 1 (resp. 0) if $\varphi$ is true (resp. false).

Two important series are $\mathbf{0}$ and $\mathbf{1}$ that are defined as $\mathbf{0}(k) \equiv 0$ and $\mathbf{1}(k) = 1$ (resp. 0) if $k = 0$ (resp. $k \neq 0$) for all integer $k$. They are the identity element for the sum and product of series respectively. The *order-of-magnitude* of $s \in \mathcal{S}_\rho$ is defined as $s^\circ \doteq \inf\{k \in \mathbb{Z} : s(k) \neq 0\}$ and $\mathbf{0}^\circ \doteq \infty$. For the subset $\underline{\mathcal{S}}_\rho$ of series such that $s^\circ > -\infty$, we have that $\langle \underline{\mathcal{S}}_\rho, +, \cdot, \mathbf{0}, \mathbf{1}\rangle$ is a field.

**Example 1:** Let $s = \sum_{k \geq 0} 2^{-k}\varepsilon^k$. Then, its multiplicative inverse is $s^{-1} = \mathbf{1} - \frac{1}{2}\varepsilon$. To check this, for $k > 0$,

$$\begin{aligned}
(s \cdot s^{-1})(-k) &= 0, \\
(s \cdot s^{-1})(0) &= s(0)s^{-1}(0) = 1, \\
(s \cdot s^{-1})(k) &= \sum_i s(i)s^{-1}(k-i) \\
&= s(k-1)s^{-1}(1) + s(k)s^{-1}(0) \\
&= -2^{-k+1}2^{-1} + 2^{-k} = 0.
\end{aligned}$$

Hence $s \cdot s^{-1} = \mathbf{1}$. □

It is not hard to show that the set $\mathcal{Q}$ of extended reals is a dense set in $\mathcal{S}_\rho$, i.e. that for any $s \in \mathcal{S}_\rho$ there exists a sequence $\langle s_n \rangle_{n \geq 0}$ from $\mathcal{Q}$ such that $s_n \to s$. In the rest of this section, we will present some general definitions needed to construct the qualitative MDP and POMDP processes.

### 3.2 A Linear Order in $\underline{\mathcal{S}}_\rho$

The construction is done in the standard way by defining the set $\mathcal{P}$ of positive elements in $\underline{\mathcal{S}}_\rho$. Let us denote with $\succeq$ the order in $\mathcal{Q}$ and let $s \in \underline{\mathcal{S}}_\rho$ be different from $\mathbf{0}$. Since $\mathcal{Q}$ is dense, there exists a sequence $\langle s_n \rangle_{n \geq 0}$ from $\mathcal{Q}$ that converges to $s$. Moreover, we can choose the sequence so that $s_n{}^\circ \leq s^\circ$ for all $n$. Then, we say that $s \in \mathcal{P}$ if and only if there exists an integer $N$ such that $s_n \succ \mathbf{0}$ for all $n > N$. The following shows that $\mathcal{P}$ is well-defined and satisfies the desired properties.

---

[4]Indeed, $\langle \mathcal{S}_\rho, \|\cdot\|_\rho \rangle$ is the $L_1(\mu_\rho)$ space with respect to the measure $\mu_\rho$ over $\mathbb{Z}$ defined by $\mu_\rho\{k\} = \rho^{-k}$. The Riesz-Fischer Theorem in Analysis asserts that this space is complete [23, Ch.11.§7].

**Theorem 1** *Let $s \in \underline{\mathcal{S}}_\rho$ be different from $\mathbf{0}$. Then, (a) $s$ is in (or not in) $\mathcal{P}$ independently of the chosen series $\langle s_n \rangle$, (b) either $s \in \mathcal{P}$ or $-s \in \mathcal{P}$, and (c) $s \in \mathcal{P}$ if and only if $s(s^\circ) > 0$.*

For $s,t \in \mathcal{S}_\rho$ we say that $s > t$ if and only if $s - t \in \mathcal{P}$, the other relations $<, \geq, \ldots$ are defined in the usual way. As $\mathcal{Q}$, the field $\underline{\mathcal{S}}_\rho$ also lacks the *least upper bound* property of the reals, i.e. that every bounded set has a least upper bound and greatest lower bound.

### 3.3 Normed Vector Spaces

An $n$-dimensional normed vector $\mathcal{S}_\rho^n$ with elements in $\mathcal{S}_\rho$ can be defined using the norm $\|J\| \doteq \sup_{i=1,\ldots,n} \|J(i)\|_\rho$. Since $\mathcal{S}_\rho$ is complete, $\mathcal{S}_\rho^n$ is also complete. A map $T : \mathcal{S}_\rho^n \to \mathcal{S}_\rho^n$ is a contraction mapping with coefficient $\alpha \in [0,1)$ if $\|TJ - TH\| \leq \alpha\|J - H\|$ for all $J,H \in \mathcal{S}_\rho^n$. If so, $T$ has a unique fix point $J^* \in \mathcal{S}_\rho^n$.

We will also deal with more general vector spaces whose elements can be thought as mappings $\mathcal{S}_\rho^X$ where $X$ is a (finite or infinite) set. The corresponding norm is $\|J\| = \sup_{x \in X} \|J(x)\|_\rho$. Thus, if $|X| = n$, $\mathcal{S}_\rho^X$ is the $n$-dimensional space $\mathcal{S}_\rho^n$ and if $|X| = \infty$, then $\mathcal{S}_\rho^X$ is infinite dimensional.

### 3.4 $\mathcal{S}_\rho$-Probability Spaces

An $\mathcal{S}_\rho$-probability space is a triplet $(\Omega, \mathcal{F}, P)$ where $\Omega$ is a *finite* set of outcomes, $\mathcal{F}$ is the set of all subsets of $\Omega$, and $P$ is a $\mathcal{S}_\rho$-valued function on $\mathcal{F}$ such that:

(a) $P(A) \geq \mathbf{0}$ for all $A \subseteq \Omega$,

(b) $P(A \cup B) = P(A) + P(B)$ for all disjoint $A,B \subseteq \Omega$,

(c) $P(\Omega) = \mathbf{1}$.

In this case we say that $P$ is a $\mathcal{S}_\rho$-probability over $\Omega$, or that $P$ is a *qualitative probability* over $\Omega$. A random variable $X$ on $(\Omega, \mathcal{F}, P)$ is a mapping $\Omega \to \mathcal{S}_\rho$, and its expected value is $EX \doteq \sum_{\omega \in \Omega} X(\omega)\,P(\{\omega\})$.

### 3.5 Kappa Rankings

A kappa ranking is a function $\kappa$ that maps subsets in $\mathcal{F}$ into the non-negative integers plus $\infty$ such that: $\kappa(\emptyset) = \infty$, $\kappa(\Omega) = 0$ and $\kappa(A \cup B) = \min\{\kappa(A), \kappa(B)\}$ for disjoint $A, B \subseteq \Omega$. Equivalently, a kappa ranking is a function from $\Omega$ to the non-negative integers plus $\infty$ such that $\kappa(\omega) = 0$ for at least one $\omega \in \Omega$. Such function is extended to $\mathcal{F}$ using the min. A kappa ranking should be thought as a ranking on worlds into degrees of *disbelief* so that values of 0,1,2,... refer to situations deemed as probable, unprobable, very unprobable, etc. Kappa rankings have a close connection with qualitative probabilities since if $P$ is a qualitative probability, then $P^\circ$ is a kappa ranking. They had been used to model order-of-magnitude approaches to decision making [25, 14, 21], and are closely related to other qualitative approaches like [9].

**Embeddings**

As we saw, a qualitative probability $P$ defines a *consistent* kappa ranking $P^\circ$. However, the other direction is not uniquely defined, i.e. given a kappa ranking $\kappa$ there are more than one qualitative probability $P$ so that $P^\circ = \kappa$. Below, we propose one embedding $\zeta : K[\Omega] \rightarrow P[\Omega]$ that maps kappa rankings over $\Omega$ to qualitative probabilities over $\Omega$ with some nice properties. This embedding will be the fundamental tool when translating kappa ranking descriptions of tasks into consistent descriptions in the language of qualitative probabilities. To define $\zeta_\kappa$, let $n_m$ be the number of $\omega$'s with rank $m$, i.e. $|\{\omega \in \Omega : \kappa(\omega) = m\}| = n_m$. Then, $\zeta_\kappa$ is defined as $\zeta_\kappa(\omega) \doteq \mathbf{0}$ if $\kappa(\omega) = \infty$, and

$$\zeta_\kappa(\omega) \doteq \frac{N_{\kappa(\omega)}}{n_{\kappa(\omega)}} \left[ \varepsilon^{\kappa(\omega)} - \sum_{j > \kappa(\omega)} [\![n_j \neq 0]\!] \varepsilon^j \right] \quad (11)$$

otherwise. The integers $N_k$ are defined as

$$N_0 \doteq 1, \quad N_k \doteq \sum_{j=0}^{k-1} [\![n_j \neq 0]\!] N_j. \quad (12)$$

**Example 2:** Let $\Omega = \{a, b, c, d, e\}$ and $\kappa$ be such that $\kappa(a) = \kappa(b) = 0$, $\kappa(c) = \kappa(d) = 1$ and $\kappa(e) = 5$. Then,

$$\zeta_\kappa(a) = \zeta_\kappa(b) = 2^{-1}(1 - \varepsilon - \varepsilon^5),$$
$$\zeta_\kappa(c) = \zeta_\kappa(d) = 2^{-1}(\varepsilon - \varepsilon^5),$$
$$\zeta_\kappa(e) = 2\varepsilon^5.$$

Clearly, $\zeta_\kappa(a) + \zeta_\kappa(b) + \zeta_\kappa(c) + \zeta_\kappa(d) + \zeta_\kappa(e) = \mathbf{1}$ so $\zeta_\kappa$ is a qualitative probability consistent with $\kappa$. □

As it can be seen in the example and definition, $\zeta_\kappa$ is some sort of "Maximum-Entropy Embedding" since it assigns equal mass to equally ranked worlds. The following theorem proofs that the embedding is consistent and some of its properties.

**Theorem 2** *Let $\kappa$ be a kappa ranking over $\Omega$. Then,*

*(a) $\zeta_\kappa$ is a qualitative proability over $\Omega$,*

*(b) $\zeta_\kappa(\omega)^\circ = \kappa(\omega)$ for all $\omega \in \Omega$, and*

*(c)* $\sup_\kappa \sum_{\omega \in \Omega} \|\zeta_\kappa(\omega)\|_\rho \rightarrow 1 \quad as \quad \rho \rightarrow \infty,$

*where the sup is over all kappa rankings over $\Omega$.*

*Proof:* Part $(b)$ follows easily from the definition. For $(a)$, we only need to show that $\sum_\omega \zeta_\kappa(\omega) = \mathbf{1}$:

$$\sum_{\omega \in \Omega} \zeta_\kappa(\omega)$$
$$= \sum_{\omega \in \Omega} [\![\kappa(\omega) \neq \infty]\!] \frac{N_{\kappa(\omega)}}{n_{\kappa(\omega)}} \left[ \varepsilon^{\kappa(\omega)} - \sum_{j > \kappa(\omega)} [\![n_j \neq 0]\!] \varepsilon^j \right]$$
$$= \sum_{k \geq 0} [\![n_k \neq 0]\!] n_k \frac{N_k}{n_k} \left[ \varepsilon^k - \sum_{j > k} [\![n_j \neq 0]\!] \varepsilon^j \right]$$
$$= \sum_{k \geq 0} [\![n_k \neq 0]\!] N_k \varepsilon^k - \sum_{j=0}^{k-1} [\![n_k \neq 0, n_j \neq 0]\!] N_j \varepsilon^k$$
$$= \sum_{k \geq 0} [\![n_k \neq 0]\!] \varepsilon^k \left[ N_k - \sum_{j=0}^{k-1} [\![n_j \neq 0]\!] N_j \right]$$
$$= [\![n_0 \neq 0]\!] \varepsilon^0 N_0 = \mathbf{1}.$$

For $(c)$, fix a kappa measure $\kappa$ over $\Omega$. Then,

$$\sum_{\omega \in \Omega} [\![\kappa(\omega) = 0]\!] \|\zeta_\kappa(\omega)\|_\rho \leq 1 + \sum_{j \geq 1} \rho^{-j} \leq 1 + \frac{1}{\rho - 1},$$

$$\sum_{\omega \in \Omega} [\![\kappa(\omega) > 0]\!] \|\zeta_\kappa(\omega)\|_\rho$$
$$\leq \sum_{k > 0} [\![n_k \neq 0]\!] N_k \left\| \varepsilon^k - \sum_{j > k} [\![n_j \neq 0]\!] \varepsilon^j \right\|_\rho$$
$$\leq \sum_{k \geq 1} [\![n_k \neq 0]\!] N_k \sum_{j \geq k} [\![n_j \neq 0]\!] \rho^{-j}$$
$$\leq 2^{|\Omega|} \sum_{k \geq 1} [\![n_k \neq 0]\!] \sum_{j \geq k} [\![n_j \neq 0]\!] \rho^{-j}$$
$$\leq 2^{|\Omega|} \sum_{k \geq 1} [\![n_k \neq 0]\!] \frac{\rho^{-k}}{1 - \rho^{-1}} \leq \frac{|\Omega| 2^{|\Omega|}}{\rho - 1}$$

since $[\![n_k \neq 0]\!] N_k \leq 2^{|\Omega|}$ (left as an exercise). Therefore,

$$\sum_{\omega \in \Omega} \|\zeta_k(\omega)\|_\rho \leq 1 + \frac{1 + |\Omega| 2^{|\Omega|}}{\rho - 1} \rightarrow 1$$

as $\rho \rightarrow \infty$ and independently of $\kappa$. □

## 4 Qualitative MDPs

A Qualitative Markov Decision Process (QMDP) is an MDP in which the quantitative information is given by qualitative probabilities and costs, i.e. a QMDP is characterized by

(QM1) a finite set of states $S = \{1, 2, \ldots, n\}$,

(QM2) a finite set of controls $U(i)$ for each $i \in S$,

(QM3) *qualitative* transition probabilities $P_{i,u}(j)$ of making a transition to $j \in S$ after applying control $u \in U(i)$ in $i \in S$, and

(QM4) a *qualitative* cost $g(i, u)$ of applying control $u \in U(i)$ in $i \in S$.

To define the cost associated to policy $\pi = (\mu_1, \mu_2, \ldots)$, consider an $N$-stage $\pi$-*trajectory* starting at state $i$ $\tau = \langle x_k \rangle_{k \geq 0}$ where $x_0 = i$ and $P_{x_k, \mu_k(x_k)}(x_{k+1}) > \mathbf{0}$. Each such trajectory $\tau$ has qualitative probability and cost given by

$$P(\tau) \doteq \prod_{k=0}^{N-1} P_{x_k, \mu_k(x_k)}(x_{k+1}), \qquad (13)$$

$$g(\tau) \doteq \sum_{k=0}^{N-1} \alpha^k g(x_k, \mu_k(x_k)). \qquad (14)$$

The *infinite horizon qualitative expected discounted cost* of applying policy $\pi$ starting at state $i$ is defined as

$$J_\pi(i) \doteq \lim_{N \to \infty} \sum_\tau P(\tau) g(\tau) = \lim_{N \to \infty} E[g(\tau)] \qquad (15)$$

where the sum is over all $N$-stage $\pi$-trajectories starting at $i$ and the expectation is with respect to the qualitative probability (13) (compare with (1) for MDPs). In general, the limit (15) is not always well-defined. However, when all costs $g(i,u)$ are in $\mathcal{Q}$ and $\alpha < 1$, then the limit exists and $J_\pi$ is well-defined. From now on, we will assume that this is the case. The optimal cost-to-go starting from state $i$, denoted by $J^*(i)$, is $J^*(i) \doteq \inf_\pi J_\pi(i)$. We would like to prove that $J^*$ is well-defined and that there exists a stationary policy $\mu^*$ such that $J^* = J_{\mu^*}$. Unfortunately, such result seems very difficult since $\underline{S}_\rho$ lacks the least upper bound property of the reals. Thus, we conform ourselves with showing the existence of optimal stationary policies and a method for computing them. That is, we need to show that the partial order $\leq$ in $\underline{S}_\rho^n$ (where $J \leq J'$ if $J(i) \leq J(i), i = 1 \ldots, n$) has unique minimum in the set $\{J_\mu : \mu \text{ a stationary policy}\}$. This result will follow if the qualitative version of the Bellman equation has unique solution. Let $T$ be the DP operator for the qualitative MDP. Then,

**Theorem 3** *If $\alpha < 1$, then there exists $\rho \geq 1$ such that $T$ is a contraction mapping.*

*Proof:* Choose $\rho$ large enough so that

$$\gamma \doteq \max_{i=1\ldots n} \sup_{u \in U(i)} \alpha \sum_{j=1}^n \|P_{i,u}(j)\|_\rho < 1,$$

which exists since $U(i)$ is finite. Let $J, H \in \mathcal{S}_\rho^n$. Then,

$$\|TJ - TH\|$$
$$= \max_{i=1\ldots n} \|(TJ)(i) - (TH)(i)\|_\rho$$
$$= \max_{i=1\ldots n} \left\| \left( \min_{u \in U(i)} g(i,a) + \alpha \sum_{j=1}^n P_{i,u}(j) J(j) \right) - \left( \min_{u \in U(i)} g(i,a) + \alpha \sum_{j=1}^n P_{i,u}(j) H(j) \right) \right\|_\rho$$

$$\leq \max_{i=1\ldots n} \alpha \sum_{j=1}^n \|P_{i,u^*}(j)\|_\rho \|J(j) - H(j)\|_\rho$$
$$\leq \|J - H\| \max_{i=1\ldots n} \alpha \sum_{j=1}^n \|P_{i,u^*}(j)\|_\rho$$
$$\leq \gamma \|J - H\|$$

where $u^*$ is the control that minimizes the minimum term in the second equality, and we used the inequality (left also as an exercise) $\|s \cdot t\|_\rho \leq \|s\|_\rho \|t\|_\rho$. □

**Corollary 4** *Assume $g(i,u) \in \mathcal{Q}$ for all $i \in S, u \in U(i)$ and $\alpha < 1$. Then, the qualitative version of the Bellman Equation (2) has unique solution $J^*$. In addition, $J^*$ can be found with Value Iteration, and the policy $\mu^*$ greedy with respect to $J^*$ is the best stationary policy.*

As a remark, note that the existence of $\rho$ proves that the Value Iteration algorithm converges, and that $\rho$ is not necessary for applying VI. Also, note that the effective discount factor is $\alpha$ since $\gamma$ can be made as close to $\alpha$ as desired (by letting $\rho \to \infty$).

### 4.1 Order-of-Magnitude Specifications

We say that a QMDP is an *order-of-magnitude specification* when the transition probabilities $P_{i,u}(j)$ are only known up to a compatible kappa ranking $\psi_{i,u}(j)$, i.e. $P_{i,u}(j)^\circ = \psi_{i,u}(j)$. In this case, we consider the QMDP that corresponds to the operator:

$$(TJ)(i) = \min_{u \in U(i)} \left\{ g(i,u) + \alpha \sum_{j=1}^n \zeta_{\psi_{i,u}}(j) J(j) \right\}$$

where $\zeta_{\psi_{i,u}}$ is the embedding of $\psi_{i,u}$. Clearly, Corollary 4 asserts that this QMDP can be solved with VI.

## 5 Qualitative POMDPs

A definition for Qualitative POMDPs (QPOMDP) can be obtained readily from the POMDP formulation by changing the POMDP formulation with:

(QP3) *qualitative* transition probabilities $p_{i,u}(j)$ of making a transition to $j$ when control $u \in U(i)$ is applied in $i \in S$,

(QP5) *qualitative* observation probabilities $p_{i,u}(o)$ of receiving observation $o \in O(i,u)$ in $i \in S$ after the application of $u \in U(i)$, and

(QP6) a *qualitative* cost $g(i,u)$ associated to $u \in U(i)$ and $i \in S$.

Similarly, we can define a qualitative version of the belief-MDP but a serious problem appears: the infiniteness of the belief-MDP thwarts a suitable choice for $\rho$ as in Theorem 3. Fortunately, we can get good results for order-of-magnitude specifications.

### 5.1 Order-of-Magnitude Specifications

A QPOMDP is said to be an *order-of-magnitude specification* if the qualitative probabilities $P_{i,u}(j)$ and $P_{i,u}(o)$ are only known up to compatible kappa rankings $\psi_{i,u}(j)$ and $\vartheta_{i,u}(o)$ respectively. In this case, we only consider *kappa belief states* which are kappa rankings over states. Thus, the transition dynamics of kappa belief states is given by the order-of-magnitude version of (6)–(9):

$$\kappa_u^o(i) \doteq \kappa_u(i) + \vartheta_{i,u}(o) - \kappa^u(o), \qquad (16)$$

$$\kappa_u(i) \doteq \min_{j=1\ldots n}\left[\kappa(j) + \psi_{j,u}(i)\right], \qquad (17)$$

$$\kappa^u(o) \doteq \min_{i=1\ldots n}\left[\kappa_u(i) + \vartheta_{i,u}(o)\right], \qquad (18)$$

$$O(k,u) \doteq \{o : k^u(o) < \infty\}. \qquad (19)$$

As the reader can check, $\kappa_u$, $\kappa_u^o$ and $\kappa^u$ are genuine kappa rankings over $S$. The qualitative belief-MDP over kappa belief states is:

(K1) A belief space $K$ of kappa measures over $S$,

(K2) a set of controls $U(\kappa) = \{u : \forall i[\kappa(i) < \infty \Rightarrow u \in U(i)]\}$ for each $\kappa \in K$, and

(K3) a cost $g(k,u) = \sum_{i=1}^{n} g(i,u)\,\zeta_\kappa(i)$ for each $u \in U(\kappa)$ and $\kappa \in K$ where $\zeta_\kappa$ is the projection of $\kappa$.

With transition probabilities given by

$$\kappa \rightsquigarrow \kappa_u^o \quad \text{with qualitative probability} \quad \zeta_{\kappa^u}(o).$$

Therefore, the Bellman equation is

$$J^*(\kappa) = \min_{u \in U(\kappa)} \left\{ g(\kappa,u) + \alpha \sum_{o \in O(\kappa,u)} \zeta_{\kappa^u}(o) J^*(\kappa_u^o) \right\} \qquad (20)$$

with the obvious DP operator. As claimed, we now prove that the Bellman equation (20) has unique solution that can be found with Value Iteration.

**Theorem 5** *If $\alpha < 1$, then there exists $\rho > 1$ such that the DP operator $T$ is a contraction mapping.*

*Proof:* Choose $\rho$ large enough such that

$$\sup_{\kappa} \sup_{u \in U(\kappa)} \alpha \sum_{o \in O(\kappa,u)} \|\zeta_{\kappa^u}(o)\|_\rho < 1$$

where the first sup is over all kappa rankings over $S$. The existence of $\rho$ is guaranteed by $\alpha < 1$ and Theorem 2-(c). Then, use a proof similar to that of Theorem 3. □

**Corollary 6** *Assume $g(i,u) \in \mathcal{Q}$ for all $i \in S, u \in U(i)$ and $\alpha < 1$. Then, the Bellman Equation (20) has unique solution $J^*$. In addition, $J^*$ can be found with Value Iteration, and the policy $\mu^*$ greedy with respect to $J^*$ is the best stationary policy.*

## 6 Computational Issues

So far, we have mainly focused the paper in the theoretical foundations of the theory and not in the computational side. However, we showed that the Value Iteration algorithm can be used to find optimal stationary policies for QMDPs and QPOMDPs. As in the standard theory, Value Iteration is one of the most general algorithms for solving sequential decision tasks, yet for certain subclasses of problems other algorithms might be more efficient. For example, problems in which the agent does not receive any feedback from the environment form an important class known as *conformant* planning problems that can be efficiently solved by performing search in belief space with standard algorithms like A* or IDA*. Such methods had proven to be powerful and successful in the standard probabilistic setting so we believe that they can be used in this setting as well [4, 5]. Another important subclass is that in which all transition and observation probabilities have the same order of magnitude 0. In such cases, the kappa belief states are just sets of states and the resulting models can be solved using model-checking or SAT based approaches to planning [7, 2, 17]. Finally, in settings in which the number of steps is bounded a priori, the qualitative planning problem can be encoded into a (qualitative) probabilistic SAT formula that can be solved by a qualitative version of the MAXPLAN planning algorithm [19].

## 7 Discussion

The work in kappa rankings was first formalized by Spohn [25] but its roots can be traced back to Adam's conditionals [1]. They have been used by Pearl and Goldszmidt to define a qualitative decision theory [22, 21, 14] and are also connected with the $\epsilon$-semantics for default reasoning [20, 12]. More recently, Giang and Shenoy presented a qualitative utility theory based in kappa rankings [13].

Another approach to qualitative MDPs and POMDPs had been recently given in terms of possibility theory [11, 24]. This approach is based on the qualitative decision criteria within the framework of possibility theory suggested in [10]. As our approach, their approach computes the value function and policy using a suitable version of the Value Iteration algorithm. However, as their example shows, the (possibilistic) cost function has not enough information for discriminating among optimal decisions. This is a fundamental departure from the standard theory of MDPs and POMDPs in which there is a one-one correspondence between stationary policies and cost functions. The reason for this difference can be better understood by considering the order-of-magnitude version of the Bellman equation:

$$J^*(i) = \min_{u \in U(i)} \max\left\{g(i,u)^\circ,\, \max_{j=1\ldots n}\left(p_{i,u}(j)^\circ + J^*(j)\right)\right\}.$$

It is not hard to see that this equation is (usually) constant for all states except the goal, so it cannot be used to discriminate among actions. This loss of information is due to the fact that order-of-magnitude quantities cannot "increase" through summations (a fact that is well-known to researchers in the field).

In summary, we have proposed a formulation for a qualitative theory of MDPs and POMDPs that is based upon a novel complete extension of the extended reals proposed by Wilson. The formal developments had been achieved using Mathematical ideas from Functional Analysis and were motivated by the necessity of taking limits. The new entities are plugged into the theory of MDPs to obtain a qualitative theory that is very close to the standard theory. This is an important difference with other approaches whose ties with the standard theory of expected utility are not as clear.

## Acknowledgements

We like to thanks the anonymous UAI reviewers for their comments. This research was supported in part by grants from NSF, ONR, AFOSR, and DoD MURI program. Blai Bonet is also supported by a USB/CONICIT fellowship from Venezuela.